%% file: article.tex
\pdfoutput=1
\def\llm{111}

\def\FrCoE{4,938}

\documentclass[11pt]{article}
\usepackage[table,dvipsnames]{xcolor}

\usepackage{acl}

\usepackage{times}
\usepackage{latexsym}

\usepackage[T1]{fontenc}
\usepackage[utf8]{inputenc}

\usepackage{tikz,pgfplots,pgfplotstable}

\usepackage{microtype}
\usepackage{hyperref}
\usepackage{booktabs}
\usepackage{graphicx}
\usepackage{multirow}
\usepackage{float}
\usepackage{adjustbox}
\usepackage{amsmath}
\usepackage{amssymb}
\usepackage{amsthm}
\usepackage{bm} 
\usepackage{mathtools, nccmath}
\usepackage{enumitem}

\usepackage{subcaption}

\usepackage[english]{babel}
\addto\extrasenglish{%

}

\usepackage{csquotes}

\usepackage{tikz}
\usetikzlibrary{shapes.misc, positioning, decorations.pathreplacing, calc}
\usepackage{tcolorbox}
\definecolor{azure}{rgb}{0.0, 0.5, 1.0}
\definecolor{tgb1}{rgb}{0.2666, 0.466, 0.666}
\definecolor{tgb2}{rgb}{0.4, 0.8, 0.9333}
\definecolor{tgb3}{rgb}{0.1333, 0.5333, 0.2}
\definecolor{tgb4}{rgb}{0.8, 0.733, 0.2666}
\definecolor{tgb5}{rgb}{0.9333, 0.4, 0.466}
\definecolor{tgb6}{rgb}{0.666, 0.2, 0.466}
\definecolor{Green}{RGB}{76, 119, 59}

\usepackage{soul}
\DeclareRobustCommand{\hlcbluefonce}[1]{{\sethlcolor{tgb1}\hl{#1}}}
\DeclareRobustCommand{\hlcyellowfonce}[1]{{\sethlcolor{tgb4}\hl{#1}}}

\newcommand{\amine}[1]{\textcolor{red}{(MA)}}

\usepackage{pgf}

\usepackage{collcell}

\usepackage{stfloats}
\usepackage{placeins}
\usepackage{expex}
\usepackage[linguistics]{forest}
\usepackage{extdash} 

\newcommand*{\MinNumber}{0.0}%
\newcommand*{\MidNumber}{60.0} %
\newcommand*{\MaxNumber}{100.0}%

\definecolor{Gray}{gray}{0.9}
\definecolor{cb-blue-green} {RGB}{ 0,  073,  073}
\definecolor{cb-green-sea}  {RGB}{ 0, 146, 146}
\definecolor{cb-rose}       {RGB}{255, 109, 182}
\definecolor{cb-salmon-pink}{RGB}{255, 182, 119}
\definecolor{cb-purple}     {RGB}{ 73,   0, 146}
\definecolor{cb-blue}       {RGB}{ 0, 109, 219}
\definecolor{cb-lilac}      {RGB}{182, 109, 255}
\definecolor{cb-blue-sky}   {RGB}{109, 182, 255}
\definecolor{cb-blue-light} {RGB}{182, 219, 255}
\definecolor{cb-burgundy}   {RGB}{146,   0,   0}
\definecolor{cb-brown}      {RGB}{146,  73,   0}
\definecolor{cb-clay}       {RGB}{219, 209,   0}
\definecolor{cb-green-lime} {RGB}{ 36, 255,  36}
\definecolor{cb-yellow}     {RGB}{255, 255, 109}
\definecolor{cb-grey}       {RGB}{233, 233, 233}

\newcommand{\ApplyGradient}[1]{%
        \ifdim #1 pt > \MidNumber pt
            \pgfmathsetmacro{\PercentColor}{max(min(100.0*(#1 - \MidNumber)/(\MaxNumber-\MidNumber),100.0),0.00)} %
            \hspace{-0.33em}\colorbox{SeaGreen!\PercentColor!Goldenrod!50}{#1}
        \else
            \pgfmathsetmacro{\PercentColor}{max(min(100.0*(\MidNumber - #1)/(\MidNumber-\MinNumber),100.0),0.00)} %
            \hspace{-0.33em}\colorbox{Red!\PercentColor!Goldenrod!50}{#1}
        \fi
}

\newcommand*{\MinNumberTwo}{-70}%
\newcommand*{\MidNumberTwo}{-30} %
\newcommand*{\MaxNumberTwo}{20}%

\newcommand{\ApplyGradienttwo}[1]{%
        \ifdim #1 pt > 0 pt
            \pgfmathsetmacro{\PercentColor}{max(min(100.0*(#1 - \MidNumberTwo)/(\MaxNumberTwo-\MidNumberTwo),100.0),0.00)} %
            \hspace{-0.33em}\colorbox{SeaGreen!\PercentColor!Goldenrod!50}{#1}
        \else
            \pgfmathsetmacro{\PercentColor}{max(min(100.0*(\MidNumberTwo - #1)/(\MidNumberTwo-\MinNumberTwo),100.0),0.00)} %
            \hspace{-0.33em}\colorbox{Red!\PercentColor!Goldenrod!50}{#1}
        \fi
}

\newcolumntype{R}{>{\collectcell\ApplyGradient}c<{\endcollectcell}}
\newcolumntype{D}{>{\collectcell\ApplyGradienttwo}c<{\endcollectcell}}

\title{Idiom Understanding as a Tool to Measure the Dialect Gap}

\author{David Beauchemin$^\dagger$, Yan Tremblay$^\dagger$, Mohamed Amine Youssef$^\dagger$ \and Richard Khoury\\
Group for Research in Artificial Intelligence of Laval University (GRAIL)\\
Université Laval, Québec, Canada \\
\texttt{david.beauchemin@ift.ulaval.ca},
\texttt{yan.tremblay.6@ulaval.ca},\\
\texttt{mohamed-amine.youssef.1@ulaval.ca}, 
\texttt{richard.khoury@ift.ulaval.ca}
}

\begin{document}
\maketitle
\begin{abstract}
The tasks of idiom understanding and dialect understanding are both well-established benchmarks in natural language processing. In this paper, we propose combining them, and using regional idioms as a test of dialect understanding. Towards this end, we propose three new benchmark datasets for the Quebec dialect of French: QFrCoRE, which contains 4,633 instances of idiomatic phrases, and QFrCoRT, which comprises 171 regional instances of idiomatic words, and a new benchmark for French Metropolitan expressions, MFrCoE, which comprises \FrCoE{} phrases.
We explain how to construct these corpora, so that our methodology can be replicated for other dialects. Our experiments with \llm{} LLMs reveal a critical disparity in dialectal competence: while models perform well on French Metropolitan, 65.77\% of them perform significantly worse on Quebec idioms, with only 9.0\% favoring the regional dialect. These results confirm that our benchmarks are a reliable tool for quantifying the dialect gap and that prestige-language proficiency does not guarantee regional dialect understanding.
\end{abstract}

\section{Introduction}
\label{sec:intro}
The task of idiom understanding is universally challenging across languages. That is because idioms, in any language, have a meaning that is often unrelated to the meanings of the individual words that compose them (e.g. \enquote{get off your high horses}), is derived from local history (e.g. \enquote{bury the hatchet}) or folklore (e.g. \enquote{a trail of breadcrumbs}), or relies on words that are no longer in use in the language (e.g. \enquote{hoist with his own petard}). But idiom understanding also relates to another natural language processing (NLP) challenge, namely dialect adaptation. For example, to describe someone who talks too much, in the USA one would say they \enquote{could talk someone's head off}, in the UK they \enquote{could talk the hind legs off a donkey}, and in Australia they \enquote{could talk underwater with a mouth full of marbles}. A language model (LM) trained in one of these dialects will have trouble making sense of the idioms from the other two. But that is exactly what happens when one dialect has more speakers and more digital data than the others and is thus more represented in training corpora, such as is the case with US compared to UK and AU English.

This problem is not exclusive to English dialects. While LLMs have proven quite proficient in standard (a.k.a. Metropolitan or Parisian) French\footnote{See \href{https://colebenchmark.org/}{https://colebenchmark.org/} for a list of French benchmarking tasks and performance results.}, little attention has been given to other French dialects. 
In this paper, we focus on idiom understanding in the Quebec dialect of French, also known as the Quebecois language. Our contributions are twofold:
\begin{enumerate}[leftmargin=*, noitemsep, topsep=0ex]
    \item We propose three \href{https://huggingface.co/datasets/graalul/QFrCoRE_QFrCoRT}{new benchmark datasets}\footnote{\href{https://huggingface.co/datasets/graalul/QFrCoRE_QFrCoRT}{https://huggingface.co/datasets/graalul/QFrCoRE\_QFrCoRT}}: 
    \begin{enumerate}[leftmargin=*, noitemsep, topsep=0ex]
        \item the \textbf{Q}uebec-\textbf{Fr}ench \textbf{Co}rpus of \textbf{R}egional \textbf{E}xpressions (QFrCoRE),
        \item the \textbf{Q}uebec-\textbf{Fr}ench \textbf{Co}rpus of \textbf{R}egional \textbf{T}erms (QFrCoRT), and
        \item the \textbf{M}etropolitan \textbf{Fr}ench \textbf{Co}rpus of \textbf{E}xpressions benchmark (MFrCoE).
    \end{enumerate}
    \item We compare LLM performances across these benchmarks, and provide a quantitative assessment of the dialect gap between French Metropolitan and Quebec French.
\end{enumerate}

The rest of this paper is organized as follows. In \autoref{sec:rel_work}, we present a review of the topics of idiom understanding and dialect understanding. We introduce our new corpora in \autoref{sec:corpora}.
Then, we present our experimental setup in \autoref{sec:experiment} and our results and discussion in \autoref{sec:results}. We then conclude and discuss our future work.

\section{Related Work}
\label{sec:rel_work} 

\subsection{Idiom and Dialect Understanding}
Studies of the challenges of idiom understanding can be found across many different languages. 
A Chinese study \cite{zheng-etal-2019-chid} found that LMs perform \enquote{much worse} than humans at idiom understanding. Two separate experiments using GPT3 in English reached the same conclusion: \citet{10.1162/tacl_a_00478} found that it understood 60\% of idioms in their dataset, well below human performance of 92\%, while \cite{coms2022miqa} found they understood 89\% of idioms in their dataset compared to 100\% understanding by humans. The disparity in performance between these two English studies demonstrates another important point: there are different levels of difficulty in understanding idioms, a realization also reached by \cite{li2024finding} with metaphors. 
Experiments in Danish showed that both ChatGPT 4 and LLaMa 3 fail to understand between a quarter to a third of idioms in that language, and have particular difficulty with culturally-specific idioms, failing to understand half of them \cite{sorensen2025danish, pedersen-etal-2025-evaluating}.

The authors of \citet{kantharuban2023quantifying} conducted a large-scale study of the performance gap across different dialect variants of the same language, which they called the \textit{dialect gap}. Using dialects of Arabic, Bengali, Finnish, Georgian, German, Malay, Mandarin, Portuguese, Spanish, Swahili, Tagalog, Tamil, and Telugu, they showed that the gap is real but highly variable across dialects. They also showed that the gap was affected by a variety of factors, including technical ones (such as the size of available training datasets), linguistic ones (dialects more lexically similar to the prestige dialect perform better), and social ones (dialects used by populations with higher GDP generally perform better). Their results demonstrate the need to study dialects individually, and not assume that observations on one language or dialect will apply uniformly to others. 
Other studies have confirmed their results by showing a performance gap for LLMs when handling various minority dialects, such as Singaporean English \cite{liangcolloquial}, Cerkno Slovenian, Chakavian Croatian, and Torlak Serbian \cite{ljubesic-etal-2024-dialect}.

Furthermore, recent work, such as \citet{kim2025memorization}, investigated  whether LLM performance relies on rote memorization or genuine conceptual reasoning in a typologically diverse multi-language setup (6 languages). Their results show that LLMs employ a hybrid approach to idiomatic understanding. This approach combines direct memorization (internal knowledge retrieval) and reasoning-based inference (compositionality and contextual cues). They also highlight a significant performance gap across languages, with LLMs performing notably better on high-resource languages (EN, DE, ZH) than on lower-resource languages (KO, AR, TR). This disparity is linked to higher memorization rates and greater model exposure in the high-resource languages.

No work has been done on the challenge of understanding idioms in minority dialects. The closest work is the study of \cite{mei-etal-2024-slang}, which proposes a causal reasoning framework to understand the meaning of idioms from context, and uses Internet slang idioms as a test case. 
While not directly comparable to other works reviewed here, their paper does demonstrate that out-of-the-box LLMs have difficulty understanding minority-dialect idioms, even in English. 

\subsection{Language Model Evaluation}
Historically, evaluation of LMs has been conducted either using mathematical metrics or benchmark corpora \cite{chang2023survey}.
The first approach relies either on task-agnostic metrics, such as perplexity \cite{jelinek1977perplexity}, which measures the quality of a model's probability distribution over word generation, or on task-specific metrics, such as the BLEU score, which evaluates a model's performance in machine translation \cite{Papineni02bleu:a}.
The second approach relies on large corpora designed for Natural Language Understanding (NLU) or natural language generation (NLG) downstream tasks. 
For example, the GLUE benchmark \cite{wang-etal-2018-glue} is used to assess a model's NLU performance on tasks such as semantic similarity, linguistic acceptability judgment and sentiment analysis. 
Likewise, GLGE \cite{liu2021glge} evaluates NLG tasks such as summarization.

Idiom understanding is evaluated using benchmark corpora that feature idiom-definition pairs, and are used to check if an LM can either give or recognize the correct definition of an idiom. \autoref{tbl:datasets} presents a list of these corpora. We do not include idiom identification datasets (which mix idioms and literal phrases and lack definitions), nor metaphor datasets (words and expressions used in a figurative sense). The Source column indicates whether the idioms and definitions were taken from a traditional dictionary, from online lexical resources, or were manually written by the authors.

\begin{table*}
    \centering
    \begin{tabular}[t]{lp{0.2\textwidth}rlc}
    \toprule
     Reference & Language (ISO-2) & Idioms & Source & Dialect? \\ 
     \midrule
     \citet{zheng-etal-2019-chid} & ZH & 3,848 & Online & No \\
     \citet{sorensen2025danish} & DA & 1,000 & Dictionary & No \\ 
     \citet{pedersen-etal-2025-evaluating} & DA & 150 & Dictionary & No \\
     \citet{10.1162/tacl_a_00478} & EN & 554 & Online & No \\
     \citet{coms2022miqa} & EN & 150 & Authors & No \\
     \citet{mei-etal-2024-slang} & EN & 408 & Online (Urban Dictionary) & Yes \\
     \citet{liu2016phrasal} & EN & 171 & Dictionary & No \\
     \citet{pershina2015idiom} & EN & 1,400 & Online (social media) & No \\
     \citet{omer2025idiom} & KU & 101 & Authors & No\\
     \citet{moussallem2018lidioms} & EN, DE, IT, PT, RU & 815 & Websites & No\\\midrule
     QFrCoRE (Ours) & \multirow{2}{*}{FR (Quebec)} & 4,633 & \multirow{3}{*}{Dictionaries and Online} & \multirow{2}{*}{Yes} \\
     QFrCoRT (Ours) & & 171  &  &  \\
     MFrCoE (Ours) & FR (Standard) & \FrCoE{} & &No\\
     \bottomrule
     \end{tabular}
     \caption{Existing idiom datasets. For each, we present the language in ISO-2 format, the number of instances (idioms), the sources of the corpus, and whether the corpus is a dialect one.}
     \label{tbl:datasets}
     \vspace{-1em}
\end{table*}

\section{QFrCoRE, QFrCoRT and MFrCoE}
\label{sec:corpora}
Our core thesis in this paper is that understanding regional idioms is an informative challenge to probe an LM's grasp of a local dialect. 
The reason is that the dialect's linguistic rules, syntax and grammar can be approximated or inferred from the prestige dialect, and thus an LM trained exclusively on that dialect can perform well on the unseen minority dialect. 
However, a dialect's idioms are unique to it, derived from the speaker population's shared culture and history, and thus cannot be easily inferred from learning about a completely different prestige population. 

\subsection{Corpora Details}
In this section, we introduce three novel evaluation corpora designed to probe LMs' understanding of idioms in the Quebecois dialect of French. 
To this end, we employ a classification task: given an expression or term and a set of definitions, the LM must select the appropriate definition.
We distinguish between two variations of this challenge: understanding either multi-word idiomatic expressions or individual idiomatic words. 
Our two datasets are QFrCoRE for idiomatic sentences and QFrCoRT for words. 
We also introduce MFrCoE, a corpus of French Metropolitan  expressions. The goal of this corpus is to assess the dialect's gap between understanding French Metropolitan  and Quebec French expressions.
All sources are publicly available with a CC-BY-NC 4.0 license.

\paragraph{QFrCoRE}
The QFrCoRE dataset comprises a set of 4,633 Quebec expressions, such as the saying \enquote{\textit{attache ta tuque avec de la broche}} (literally: fasten your toque with wire), which means that one should brace oneself for something about to happen, equivalent to the French expression \enquote{\textit{attache ta ceinture}} (literally: buckle your seatbelt). 
Its primary sources are the \enquote{Dictionnaire des expressions québécoises} \cite{desruisseaux2009dictionnaire}, supplemented by entries from the \enquote{Dictionnaire des proverbes, dictons et adages québécois} \cite{desruisseaux2008proverbes} and the \href{https://canada-media.ca/expressions-quebecoises}{Canada-Media} online portal.

\paragraph{QFrCoRT}
The QFrCoRT dataset comprises a set of 171 Quebecois words, such as the term \enquote{\textit{Tiguidou!}}, which means that something went extremely well.
The terms were scraped and deduplicated from five online collections of Quebec regional language (see \autoref{an:sources}).

\paragraph{MFrCoE}
The MFrCoE dataset comprises a set of \FrCoE{} French Metropolitan  expressions, and is intended as an equivalent of QFrCoRE in the prestige dialect. 
Its primary source is \enquote{Les 1001 expressions préférées des Français} \cite{planelles20191001}, supplemented by entries from the same dictionary online source, \href{https://www.expressio.fr/toutes-les-expressions/}{Expressio}.
\subsection{Dataset Creation Methodology}
\label{sec:methodology}

\subsubsection{Data Collection}

\paragraph{QFrCoRE}
The corpus was created by scanning the two dictionaries and manually extracting expressions from the online portal.
Then, we use an online OCR solution to extract all the dictionaries' content using out-of-the-box Azure OCR AI model through their \enquote{Document Intelligence} solutions \citep{azureaidoc}.
Second, we manually and semi-manually curated the content by using regular expressions and manual manipulation to clean the expressions and definitions.
Finally, duplicates were manually removed from the corpus. 

\paragraph{QFrCoRT}
The corpus was created by manually extracting terms and their corresponding definitions from a curated list of websites and dictionaries (see \autoref{an:sources}). Anglicisms (i.e. English words or expressions borrowed into French) and duplicates were then manually removed from the corpus. 
The curated list of websites and dictionaries was created from multiple manual Web-indexed searches using the following search keywords: \enquote{\textit{expressions québécoises}}, \enquote{\textit{expression [québec/saguenay/beauce/montréal/mauricie/abitibi /gaspésie]}}, and \enquote{\textit{termes québec}}.

\paragraph{MFrCoE}
The corpus was created by manually extracting expressions from the dictionary and online source and manually cleaning each instance.

\subsubsection{Distractor Generation}
Following \citet{sorensen2025danish} and \citet{coms2022miqa}, we opted to create a multiple-choice idiom understanding evaluation.  
For each idiom, nine distractors (false definitions) are generated using a state-of-the-art (SOTA) LLM, \texttt{GPT-4o-mini} with default parameters.
As shown in \autoref{fig:gen_prompt}, the prompts we use are designed to generate distractors that are semantically plausible but incorrect. This ensures that picking the correct answer requires an understanding of the idiom rather than superficial keyword matching. 
Furthermore, we validate that the distractors are sufficiently different from the correct definition to be unambiguously incorrect, first automatically using similarity metrics, and then through manual review.
First, we compare each distractor to the correct definition using a weighted average of BERTScore \citep{zhang2019bertscore}, ROUGE \citep{lin-2004-rouge} and BLEU \citep{Papineni02bleu:a} using respective weights of 0.470, 0.176 and 0.353.
Weights have been selected through trial and error, yielding relevant distractors that are neither too similar nor too irrelevant.
A distractor that yields a score higher than a threshold of 0.45 is considered too similar to the true definition and is rejected.
The threshold value has been selected through trial and error.
This triggers a new generation, but using \texttt{GPT-4.1} instead of \texttt{GPT-4o-mini}. 
In addition, each iteration with rejected distractors increases the generation temperature by 0.1, up to a maximum of 1.6.
We present in \autoref{tab:examples} translated examples, their metrics for accepted and rejected distractors, distractors with our algorithm's maximum temperature, and the reasons for acceptance or rejection.
Second, we manually validate that the final distractors are sufficiently related and relevant.
Finally, the true definition is randomly positioned among the distractors, so that each of the 10 answers has roughly equal probability of being the correct one. 

\begin{figure*}[ht!]
    \begin{subfigure}{0.49\linewidth}
        \centering
        \begin{tikzpicture}[scale=1, every node/.style={transform shape}]
        \node[rectangle, rounded corners, draw=tgb4, fill=tgb4, text width=0.8\linewidth, align=left, inner sep=1ex] (prompt) {
        For the [Quebec/French] term: \enquote{\{term\}}, the correct definition is \enquote{\{definition\}}. Generate a different definition, believable, but wrong.
        };
        \end{tikzpicture}
        \caption{QFrCoRT/MFrCoE}
        \label{fig:QFrCoRT_gen}
    \end{subfigure}
    \begin{subfigure}{0.49\linewidth}
        \centering
        \begin{tikzpicture}[scale=1, every node/.style={transform shape}]
        \node[rectangle, rounded corners, draw=tgb4, fill=tgb4, text width=0.8\linewidth, align=left, inner sep=1ex] (prompt) {
        For the Quebec expression: \enquote{\{expression\}}, the correct definition is \enquote{\{definition\}}. Generate a different definition, believable, but wrong.
        };
        \end{tikzpicture}
        \caption{QFrCoRE}
        \label{fig:QFrCoRE_gen}
    \end{subfigure}
    \caption{
    The prompts used for generating distractors, we use \enquote{Quebec} for QFrCoRT, and \enquote{French} for MFrCoE.
    }
    \label{fig:gen_prompt}
\end{figure*}

\begin{table*}[ht]
    \centering
    \resizebox{0.99\textwidth}{!}{%
    \begin{tabular}{p{2.5cm} p{7.5cm} c c c cc}
        \toprule
        Definition & Distractor &\begin{tabular}[c]{@{}c@{}}BERTScore\\(\%)\end{tabular} & \begin{tabular}[c]{@{}c@{}} ROUGE\\(\%)\end{tabular}  & \begin{tabular}[c]{@{}c@{}}BLEU\\(\%)\end{tabular} & \begin{tabular}[c]{@{}c@{}}Agg. Score\\(\%) (AS)\end{tabular} & \begin{tabular}[c]{@{}c@{}}Status: Reason\end{tabular}\\
        \midrule
        & To be seized with great anger or fury & 79.35 & 47.06 & 0.00 & 45.64 &Rejected: AS $> 45.00$\\ 
        \multirow{3}{*}{\parbox{2.5cm}{To be angry, furious}}  & Being in the process of cooking a dish made from boar meat & 66.41  & 25.00  & 0.00  & 35.66  &Accepted: AS $< 45.00$\\
        & To be lost in thought, to think intensely about a problem& 67.70 & 9.09 & 0.00 & 25.60 &Accept. (T $=1.6$): AS $< 45.00$\\
        \midrule
        & To act like a show-off, especially in front of a girl & 87.98 & 58.82 & 40.35 & 66.03 & Rejected: AS $> 45.00$\\
        \multirow{3}{*}{\parbox{2.5cm}{To show off, especially in front of a girl}} & To show off means to prepare for a sports competition with enthusiasm & 72.76  & 28.57  & 0.00  & 39.28  &Accepted: AS $< 45.00$\\
         & Tending to your vegetable garden by preparing the soil for summer & 69.25 & 9.09 & 0.00 & 26.11 & Accept. (T$=1.6$): AS $< 45.00$\\
        \midrule\midrule
         \multirow{3}{*}{\parbox{2.5cm}{Good luck in French, equivalent to the expression \enquote{break a leg}}}& French expression used to wish someone good luck, similar to the expression \enquote{break a leg} & 82.07 & 59.26 & 19.35 & 55.91 &Rejected: AS $> 45.00$\\
          & A term used to express great joy or unexpected success & 63.45  & 0.00  & 0.00  & 29.86 &Accepted: AS $< 45.00$ \\
        & Ancient prayer recited before beginning an important meeting & 66.14 & 0.00 & 0.00 & 22.05  & Accept. (T$=1.6$): AS $< 45.00$ \\
        \midrule
        & This expression means \enquote{not at all} & 89.68 & 36.36 & 48.62 & 65.71 &Rejected: AS $> 45.00$\\
        \multirow{3}{*}{\parbox{2.5cm}{This expression means \enquote{not at all}}} & This expression refers to a type of traditional Quebec dance & 72.68  & 22.22  & 0.00  & 38.12 &Accepted: AS $< 45.00$ \\
        & This expression means \enquote{wait a minute} & 88.29 & 50.00 & 0.00 & 39.97 & Accept. (T$=1.6$): AS $< 45.00$ \\
        \bottomrule
        \end{tabular}%
    }
    \caption{Examples of translated generated distractors were evaluated against the correct definition using three metrics and an aggregated score (AS). Candidates with an $AS > 45.00$ are rejected due to excessive similarity to the reference. The upper part of the table shows distractors for QFrCoRE, while the lower part is for QFrCoRT. \enquote{T$=1.6$} means that the distractor was generated with our algorithm's maximum temperature value.}
    \label{tab:examples}
    \vspace{-0.5em}
\end{table*}


\subsection{Corpora Statistics}
\autoref{tab:stats} presents statistics of the corpora, where the lexical richness corresponds to the type-token ratio (TTR), i.e. the ratio of unique words to total words in an instance, without removing stop words or normalizing them \cite{van2007comparing}.
Our corpus statistics indicate that QFrCoRE is a large-scale corpus with a vast vocabulary (25,181 words) and longer phrasings in both instances (avg. 5.01 words) and answers (avg. 9.69 words). The MFrCoE corpus shares many characteristics with QFrCoRE, including a large vocabulary (21,452 words), relatively long instances (avg. 3.86 words), and answers (avg. 10.41 words). 
Conversely, QFrCoRT is much smaller, with a vocabulary of only 2,855 words and very short instances, which is expected since it deals with individual idiomatic words. 
Finally, all three corpora have a similar average lexical richness.

\begin{table}
    \centering
    \resizebox{0.49\textwidth}{!}{%
    \begin{tabular}{lccc}
    \toprule
     & QFrCoRE & QFrCoRT & MFrCoE\\
    \midrule
    Avg. instance WC & 5.01 & 1.00 &3.86 \\
    Avg. answers WC & 9.69 & 7.08 & 10.41\\
    Vocabulary size & 25,181 & 2,855 & 21,452\\
    Avg. lexical richness & 0.65 & 0.73 & 0.63\\
    \bottomrule
    \end{tabular}%
    }
    \caption{Statistics of all three corpora, where \enquote{WC} stands for \enquote{word count}.}
    \label{tab:stats}
\end{table}

\section{Experimental Setup}
\label{sec:experiment} 
In this section, we present our experimental setup.
First, we present our evaluation settings in \autoref{sec:eval} and our \llm{} benchmarked models and baselines in \autoref{sec:models}.

\subsection{Evaluation Settings}
\label{sec:eval}
We evaluate a diverse set of French and multilingual LLMs in a zero-shot setting, with no task-specific fine-tuning or adaptation. The model is expected to produce an appropriate answer based solely on its pretrained capabilities. 

Each task is presented to the LLM using a prompt.
Our prompts were inspired by \citet{aparovich2025belarusianglue} and by prompt engineering best practices \citep{marvin2023prompt, ye2024prompt, li2024generation, bjerg2024tips}. 
Each prompt is composed of a system message providing task instructions and the idiom to be understood, followed by a user message with answer options and a placeholder for the LLM's choice.
The prompts are written in French, but we present translated versions in \autoref{fig:ZeroShotPrompts}.

\paragraph{Dialect Gap}
To assess the dialect gap between Quebec-French and French Metropolitan, we compare each LLM's result over QFrCoRE against its result on MFrCoE using a Z-test for statistical significance \cite{lawley1938generalization}. 
Our null hypothesis is that the pair of accuracies are equal, meaning that Z-test values outside the interval $[-3.290527, 3.290527]$ allow us to reject the hypothesis with $\alpha = 0.001$ (i.e. not a significant difference between the two benchmarks). 
A positive value means that French Metropolitan  has a significantly better performance than Quebec French, and a negative value means the opposite.

\begin{figure*}[ht!]
   \centering
   \begin{subfigure}{0.5\linewidth}
       \centering
       \begin{tikzpicture}[scale=1, every node/.style={transform shape}]
       \node[rectangle, rounded corners, draw=tgb1, fill=tgb1!80, text width=0.8\linewidth, align=left, inner sep=1ex] (prompt) {
       $\ll$system$\gg$ What does the [Quebec/French] \enquote{\{\textbf{expression}\}} mean? 
       Answer only with the index (starting at zero) of the correct definition. For example, if the third one is correct, answer 2.
       };
       \node[rectangle, rounded corners, draw=tgb4, fill=tgb4, below=0.1cm of prompt,  text width=0.8\linewidth, align=left, inner sep=1ex] (input) {
       $\ll$user$\gg$ 
       Here is a list of possible definitions:\{definitions\}\\
       The answer is: \{input\}.
       };
       \end{tikzpicture}
       \caption{QFrCoRE/MFrCoE}
       \label{fig:QFrCoRE}
   \end{subfigure}
   \begin{subfigure}{0.49\linewidth}
       \centering
       \begin{tikzpicture}[scale=1, every node/.style={transform shape}]
       \node[rectangle, rounded corners, draw=tgb1, fill=tgb1!80, text width=0.8\linewidth, align=left, inner sep=1ex] (prompt) {
       $\ll$system$\gg$ What does the Quebec \enquote{\{\textbf{term}\}} mean? Answer only with the index (starting at zero) of the correct definition. For example, if the third one is correct, answer 2.
       };
       \node[rectangle, rounded corners, draw=tgb4, fill=tgb4, below=0.1cm of prompt, text width=0.8\linewidth, align=left, inner sep=1ex] (input) {
       $\ll$user$\gg$ 
       Here is a list of possible definitions:\{definitions\}\\
       The answer is: \{input\}.
       };
       \end{tikzpicture}
       \caption{QFrCoRT}
       \label{fig:QFrCoRT}
   \end{subfigure}
   \caption{
   The translated prompt templates used for the zero-shot evaluation of our two benchmarks. 
   Each prompt consists of a system message providing the instruction and a user message containing the \texttt{{input}} placeholder for the data instance.
   \hlcbluefonce{Blue} boxes contain the task instructions. 
   \hlcyellowfonce{Yellow} boxes contain the prefix for the model to continue. Texts in \enquote{\texttt{$\ll\gg$}} are role-tags to be fed to the model; we use \enquote{Quebec} for QFrCoRT, and \enquote{French} for MFrCoE.}
   \label{fig:ZeroShotPrompts}
   \vspace{-1em}
\end{figure*}

\subsection{Models}
\label{sec:models}
\subsubsection{Baseline}
As a baseline, we use a \texttt{Random} selection algorithm, which picks one of the 10 possible answers at random. 
We use the seed $42$ to facilitate the reproducibility of our results.

\subsubsection{LLM}
To ensure a thorough and representative analysis of the current LM landscape, we selected \llm{} LLMs to cover five aspects of LLM specifications:

\begin{enumerate}[leftmargin=*, noitemsep, topsep=0ex]
    \item \textbf{Variety of Access Paradigms:} We included both proprietary LM accessible via an API (e.g. OpenAI) and open-source models (e.g. Llama). 
    This enables us to compare the performance of commercial offerings with models that support full customization and local deployment. 
    \item \textbf{Variety in Size:} The selected models span a large range of parameter counts, from smaller models under 1 billion parameters to the largest proprietary models available as of mid-2025.
    \item \textbf{Variety in Capability:} We intentionally included models marketed as having advanced \enquote{reasoning} capabilities (e.g. Deepthink) to assess if this specialization translates to better performance on our knowledge-intensive task.
    \item \textbf{Model Specialization:} We included models fine-tuned in French  (e.g. Chocolatine), to test whether this linguistic specialization provides an advantage.
    \item \textbf{Instruction-Tuning}: We included models that have been instruction-tuned to compare against their base model counterpart (e.g. \texttt{Apertus-8B-\textbf{it}} vs. \texttt{Apertus-8B}) .
\end{enumerate}

To select the LMs, we leverage two leaderboards: the \href{https://lmarena.ai/leaderboard/text}{Text Arena Leaderboard} and the \href{https://huggingface.co/spaces/open-llm-leaderboard/open_llm_leaderboard#/}{Open LLM Leaderboard} \citep{open-llm-leaderboard-v2}.
We present our selected models and details in \autoref{an:results}, and our hardware and private LLM budget in \autoref{an:hardware}.

\section{Results and Discussion}
\label{sec:results}

We present in \autoref{fig:accuracy} a visual representation of the accuracy of each model tested on the QFrCoRT (x-axis) and QFrCoRE (y-axis) benchmarks. Complete results are included in \autoref{an:results}. 
To simplify the analysis, we separated the models into three groups based on their performance. 
In \autoref{fig:accuracy}, models in red have a performance that is lower than our random-selection baseline (marked by the black dashed lines) on at least one of the two benchmarks. 
This cluster contains 40 models, or over a third of the models we tested. 
At the other end, we can see a cluster of high-performing models in the top-right corner of the graphic. This cluster, marked in green, comprises 27 models that achieve greater than 80\% accuracy on both benchmarks. These are mostly variations of Claude, GPT, Gemini, Grok, o1/o3 and DeepSeek.
Finally, the remaining 45 models, marked in blue in the figure, exhibit intermediate performance.

\begin{figure*}
    \centering
    \pgfplotstableread{data0 blue
12.87	10.64
12.28	11.33
12.87	11.74
12.87	12.3
17.54	12.67
32.16	12.82
23.39	13.1
18.13	13.4
16.37	13.73
15.2	14.7
18.13	14.94
19.88	15.43
13.45	15.73
15.2	15.95
17.54	17.35
24.56	17.61
36.84	18.33
16.37	23.74
22.81	24.45
15.2	27.78
53.22	31.04
35.67	32.18
31.58	32.87
29.82	33.13
28.07	33.48
54.39	34.08
45.61	34.79
38.6	39.02
40.35	41.27
67.84	44.57
56.14	47.2
53.22	48.5
74.85	53.87
50.88	57.39
70.18	64.23
76.61	66.52
70.76	67.8
77.78	70.49
90.64	73.88
79.53	75.74
87.72	76.97
86.55	77.25
92.4	77.36
92.4	79.43
91.23	79.47   
    }\dataZ
    \pgfplotstableread{data1 red
5.85	2.18
6.43	3.93
2.92	3.95
12.28	5.42
4.68	6.15
8.77	6.65
8.19	6.76
5.85	8.29
12.87	8.72
8.19	9.26
7.6	9.37
9.36	9.37
9.94	9.37
11.11	9.41
12.87	9.41
11.7	9.43
12.87	9.48
13.45	9.56
10.53	9.61
11.11	9.67
12.87	9.67
9.94	9.76
11.11	9.76
14.04	9.76
9.36	9.78
12.87	9.82
11.7	9.86
10.53	9.91
12.87	9.99
7.6	10.12
9.36	10.25
11.7	10.32
11.11	10.49
9.36	10.58
8.19	10.99
10.53	11.33
7.6	11.37
5.26	12.84
12.28	15.93
9.36	20.14
    }\dataO
    \pgfplotstableread{data2 green
    85.07	80.71
    90.64	82.5
    88.89	82.54
    92.4	82.54
    92.4	82.8
    85.38	82.82
    87.72	82.99
    89.47	83.14
    92.4	83.92
    93.57	83.94
    90.64	84.03
    90.64	84.8
    91.23	84.98
    95.91	85.41
    88.45	85.65
    95.32	85.84
    95.91	86.01
    94.15	86.73
    92.98	87.29
    95.91	87.8
    95.32	87.93
    95.91	90.09
    96.49	90.68
    97.66	91.75
    97.66	93.4
    97.66	93.46
    99.42	95.38
    }\dataT

    \begin{tikzpicture}
        \begin{axis}[
        xmin=0,
        xmax=100,
        ymin=0,
        ymax=100,
        xlabel=QFrCoRT Accuracy (\%),
        ylabel=QFrCoRE Accuracy (\%),
        xtick pos=left, 
        ytick pos=left, 
        grid=none,
        legend pos=north west,
        ]
        \addplot+ [mark=*, only marks, mark options={solid, fill=blue!60!black, opacity=0.5}] table[y=blue] {\dataZ};
        \addplot+ [mark=*, only marks, mark options={solid, fill=red!60!black, opacity=0.5}] table[y=red] {\dataO};
        \addplot+ [mark=*, only marks, mark options={solid, fill=green!60!black, opacity=0.5}] table[y=green] {\dataT};
        \addplot[mark=none, black, dashed, domain=0:100, line width=0.25mm] {9.93}; 
        \addplot[mark=none, black, dashed, line width=0.25mm] coordinates {(12.28,0)(12.28,100)}; 
        \end{axis}
    \end{tikzpicture}
    
    \caption{Accuracy plot of all \llm{} models tested, we present performance on QFrCoRT (x-axis) and QFrCoRE (y-axis). \textcolor{black}{\textbf{Black}} dashed lines are our \texttt{Random} baseline scores. \textcolor{red!60!black}{\textbf{Red}} dots are models that performed poorer than the baseline on one of the corpora, \textcolor{green!60!black}{\textbf{green}} dots are models that performed better than 80\% on both corpora, while \textcolor{blue!60!black}{\textbf{blue}} dots are those that do not fit in the two other performance classes.}
    \label{fig:accuracy}
    \vspace{-1em}
\end{figure*}

\subsection{A Challenge of Lexical Knowledge, Not Syntactic Complexity}
An important initial observation is that the scatter plot is almost linear, meaning that models have similar performances in QFrCoRT and QFrCoRE. Numerically, the average difference in score for a model between the two benchmarks is 5.5\%. 
This indicates that both benchmarks are equivalent in the language skills they test and in their difficulty level. 
This near-linear performance correlation is particularly consequential when compared to the corpus statistics in \autoref{tab:stats}. Despite QFrCoRE having a vocabulary nearly nine times larger and instances that are five times longer than those in QFrCoRT, models do not find it significantly more difficult. 
This suggests that the primary challenge lies not in syntactic complexity or general lexical breadth, but in the specialized, regional nature of the vocabulary itself. 
For current LLMs, understanding a single, culturally-embedded term is as difficult as deciphering a multi-word idiomatic phrase. 
This reinforces that the dialect gap is fundamentally a problem of cultural, not of general linguistic processing ability.

\subsection{Limited Impact of Size, Reasoning, Instruction Tuning, and Model Family}
\label{sec:irrelevant}
Our results show only a weak correlation between model size and performance on our benchmark. While, generally speaking, a larger model performs better, there is considerable variability, and many exceptions exist to this rule. For example, our seven models with 32B parameters have QFrCoRE accuracies that vary between 3.95\% and 53.87\%. The LLaMa-3.2 model with 3B parameters outperforms several 8B models and even the OLMo 32B models. 

Likewise, our results show no benefits related to the reasoning abilities of models. The 43 reasoning models we tested are distributed almost evenly in each of the three performance groups. This makes sense, since our benchmark tests knowledge and not reasoning. If an LLM does not know the meaning of an idiom, it cannot reason to figure it out. 

Instruction tuning also has no benefits in this task. The 29 instruction-tuned models are split almost evenly between the low-performance and intermediate-performance zones. Comparing the accuracy of the 21 models for which we have base and instruction-tuned versions reveals a very minor advantage for the latter one, on average only 2.3\% on QFrCoRE. This may be due to the fact that the instructions for our task were fairly simple, as seen in \autoref{fig:ZeroShotPrompts}, so instruction tuning had no impact.

Finally, our results reveal significant performance variance within the same model family. 
For instance, the DeepSeek models have QFrCoRE accuracies that range from 3.93\% to 84.98\%, and Qwen models range from 9.41\% to 75.74\%. 
This suggests that learning dialect-specific knowledge is not dependent on the architecture of the LLM.

\subsection{Impact of Language Training}
Perhaps the most striking result we found is that fine-tuning a model in French Metropolitan  does not lead to good performances on our benchmarks. Indeed, of the eight French models we tested, four are part of the low-performing group which performed worse than our random-guess baseline on at least one benchmarks, and the remaining three have intermediate performances. None are part of the top-performing group. 

The cause of this is likely the data that was used to fine-tune these models. Chocolatine and French-Alpaca are trained using English texts machine-translated into prestige-dialect French, while Lucie was trained with a dataset collected in France by its France-based makers. Thus, none of these models were exposed to data written in the Quebecois dialect during their training. The Croissant LLM is an exception, having been trained on a multinational French dataset that does include Quebec French documents, but that still heavily favors France data \cite{faysse2024croissantllm}. 

This result is thus another symptom of the dialect gap documented by \cite{kantharuban2023quantifying}. It also confirms the core hypothesis that underlies our work: the idiom understanding task is a good measure of dialect understanding. 

\subsection{Impact of Access Paradigm}
The most important feature of model performance is its access paradigm. Indeed, all low-performance models and 84\% of intermediate-performance models are open-source, while 85\% of high-performance models are proprietary. The average performance of proprietary models on QFrCoRE is 83\% and on QFrCoRT is 91\%, while open-source models, excluding the low-performance ones, only achieve average performances of 35\% and 40\% on QFrCoRE and QFrCoRT respectively. 

The larger sizes of proprietary models partly explain this result. Indeed, due to hardware limitations, we were unable to run open-source models of comparable sizes to those of proprietary models accessible through APIs. However, our previous results in \autoref{sec:irrelevant} have shown that neither size, model family, nor advanced capabilities like reasoning and instruction-tuning are predictors of high performance. We believe the real difference stems from training data: these larger proprietary models are trained on much larger datasets collected from varied online sources. If these sources include Quebec content, such as websites, novels, news articles, or Wikipedia pages, then the models would have been exposed to the Quebecois language and its idioms. Moreover, if these sources include the same publicly-available resources we collected our idioms from, then these strong results are due to data contamination. Ultimately, since the datasets used to train these models are not public, it is impossible for us to know.

\subsection{Dialect Gap Assessment}
To assess the dialect gap, we compare model performance on QFrCoRE with its French Metropolitan  counterpart MFrCoE. 
To do so, as stated in \autoref{sec:eval}, we conduct a Z-test, where our null hypothesis is that the pair of accuracies are equal.
We present in \autoref{fig:ztest} a visualization of the Z-test results, where the diagonal line represents equal performance across both dialects and the dotted lines represent the statistical significance bands.
A visual inspection reveals a distinct dialect gap below the linguistic parity, illustrating a systematic bias toward the prestige dialect.
This asymmetry is not uniform across model capabilities; it is particularly pronounced in the intermediate performance cluster (\textcolor{blue!60!black}{\textbf{blue}}). Here, models demonstrate strong competency in French Metropolitan  (x-axis) but fail to transfer this knowledge to the Quebec dialect (y-axis). 
Even among the high-performing models (\textcolor{green!60!black}{\textbf{green}}), which cluster tightly in the upper-right quadrant, the majority remain below the diagonal, indicating that the dialect gap persists, albeit to a lesser degree, even in the most capable systems. 
Conversely, most of the low-performing models (\textcolor{red!60!black}{\textbf{red}}) cluster near the random baseline on both axes, demonstrating a lack of fundamental French understanding regardless of dialect.

Moreover, as shown in \autoref{tab:z_test_stats}, which quantifies this disparity, the results demonstrate a clear performance bias in favour of the prestige dialect: 65.77\% of the evaluated models performed significantly better on the MFrCoE. Meanwhile, 25.23\% of models showed no statistically significant difference between the two dialects, and only 9.01\% performed significantly better on QFrCoRE.
These statistics confirm that for the vast majority of current LLMs, French Metropolitan  expressions are more readily understood than regional Quebec idioms. 
Moreover, almost all models that perform better on QFrCoRE are below the 20\% accuracy line, meaning they only perform better on QFrCoRE because their low performance on that benchmark is better than their abysmal performance on MFrCoE. Only one model, \texttt{Qwen3-235b-a22b-thinking-2507}, showed an actual good performance on the benchmark and a stronger performance on QFrCoRE (64.00\%) than on MFrCoE (55.67\%).

\begin{figure*}
    \centering
    \resizebox{0.49\textwidth}{!}{%
    \begin{tikzpicture}
    \begin{axis}[
        width=12cm, height=12cm,
        title={Model Accuracy Comparison between MFrCoE and QFrCoRE},
        xlabel={MFrCoE Accuracy (\%)},
        ylabel={QFrCoRE Accuracy (\%)},
        xmin=0.00, xmax=100.00,
        ymin=0.00, ymax=100.00,
        xtick pos=left, 
        ytick pos=left, 
        grid=none,
        legend pos=north west,
        scatter/classes={
            green_dot={mark=*, draw=green!60!black, fill=green!60!black, opacity=0.5},
            red_dot={mark=*, draw=red!60!black, fill=red!60!black, opacity=0.5},
            blue_dot={mark=*, draw=blue!60!black, fill=blue!60!black, opacity=0.5}
        }
    ]
    
    \addplot [black, solid, domain=0:100] {x};
    
    \draw [black!70, dashed, line width=1pt] (axis cs:10.40907249898744, 0) -- (axis cs:10.40907249898744, 100);
    \draw [black!70, dashed, line width=1pt] (axis cs:0, 9.92877185409022) -- (axis cs:100, 9.92877185409022);
    
    \addplot [gray, dotted, line width=2pt] table [x=x, y=y_upper, col sep=comma] {data/pgf_sig_curves.csv};
    \addplot [gray, dotted, line width=2pt] table [x=x, y=y_lower, col sep=comma] {data/pgf_sig_curves.csv};
    
    \addplot [scatter, only marks, scatter src=explicit symbolic] 
        table [x=acc_frcoe, y=acc_qfrcore, meta=tex_class, col sep=comma] {data/pgf_scatter_data.csv};
    
    \end{axis}
    \end{tikzpicture}%
    }
    \caption{Accuracy comparison between MFrCoE and QFrCoRE. The black dash-dotted line represents equal performance. The gray dotted lines represent the statistical significance interval using a Z-test ($\alpha=0.001$). \textcolor{black}{\textbf{Black}} dashed lines are our \texttt{Random} baseline scores. 
    \textcolor{red!60!black}{\textbf{Red}}, \textcolor{green!60!black}{\textbf{green}}, and \textcolor{blue!60!black}{\textbf{blue}} correspond to the model performances from \autoref{fig:accuracy}.}
    \label{fig:ztest}
    \vspace{-1em}
\end{figure*}

\begin{table}[ht]
    \centering
    \begin{tabular}{lr}
        \toprule
        & \textbf{Count} \\
        \midrule
        Significantly better on MFrCoE & 73 \small{(65.77\%)} \\
        Significantly better on QFrCoRE & 10 \small{(9.01\%)} \\
         No significant difference & 28 \small{(25.23\%)} \\
        \bottomrule
    \end{tabular}\
    \caption{Summary of Z-test statistical comparison between MFrCoE and QFrCoRE with $\alpha=0.001$.}
    \label{tab:z_test_stats}
    \vspace{-1em}
\end{table}

\subsection{Societal Implications}
The results of \autoref{fig:accuracy} and \autoref{fig:ztest} have some important societal implications. Indeed, they show that, when it comes to interacting with LLMs, speakers of regional dialects are severely disadvantaged. If they wish to use their dialect and be well understood by the LLM-based application, they must use a proprietary LLM. This comes with two major drawbacks. Firstly, these LLMs are expensive to use, so dialect users will end up paying a hefty price to interact in their language. And secondly, they can only be accessed by sending one's data and prompts to the company through an API, which entails losing control of one's data: the company may take the data and use it for its own purposes. This drawback also precludes using these LLMs in applications with sensitive or restricted data, such as in the medical domain. The alternative is to use an open-source LLM, which can be executed freely and locally without transferring data to a third party. But if the users interact with these LLMs in their dialect, the performance of the tools drops, catastrophically in most cases. To maintain good performances, the users will need to forego their dialect and adopt the prestige dialect of the language. This is a form of AI colonization. 

\section{Conclusion and Future Work}
\label{sec:conclusion}
In this paper, we introduced the idea of regional idiom understanding as a benchmark for dialect understanding. We validated this approach by constructing three new datasets: QFrCoRE and QFrCoRT for the Quebec dialect, and MFrCoE for French Metropolitan expressions. Our extensive evaluation of \llm{} LLMs highlights a severe deficiency in current models. Over 40\% of models perform worse than random guessing on Quebec idioms, indicating that their training on prestige data actively misleads them via negative transfer.

Furthermore, our comparative analysis with the French Metropolitan  MFrCoE quantifies the dialect gap. We observed that 65.77\% of the models performed significantly better on French Metropolitan, while only 9.01\% favoured the regional dialect. 
Strikingly, even fine-tuned French models failed to bridge this gap, demonstrating that specialization in the prestige dialect does not translate to regional linguistic competence. 
The gap is most prominent in open-source models with intermediate capabilities, which have mastered the syntax of the language but lack the specific cultural-lexical knowledge required for regional idioms.

This work establishes a methodology for idiom-based dialect benchmarking. Our next steps involve expanding this comparative framework to other French dialects beyond Quebec (e.g. Swiss) to map the performance gap across the Francophonie. We also aim to obtain human evaluations to establish a \enquote{human dialect gap} baseline, allowing us to distinguish between the natural difficulty of regional idioms and the artificial limitations of LLM training data.

\section*{Limitations}
While we believe QFrCoRE, QFrCoRT and MFrCoE provide a valuable corpus for evaluating Quebec-French and French idiom understanding, we recognize several limitations in their construction and coverage that open paths for future enhancement.

\paragraph{AI-Generated Distractors}
A limitation stems from the fact that the distractors for both corpora were generated by an LLM. 
This process may introduce subtle, systemic patterns or \enquote{AI-generated artifacts} into the distractors. 
Consequently, the benchmark might inadvertently test a model's ability to distinguish AI-generated text from human-written text rather than its actual understanding of the Quebecois expressions. 
Models from the same family as the generator (e.g. other GPT variants) could be particularly competent at identifying these artifacts, potentially leading to inflated performance scores that do not reflect actual linguistic competency \citep{balepur-etal-2024-artifacts}
. 
\paragraph{Lexical and Dialectal Coverage}
Although QFrCoRE contains 4,633 expressions, QFrCoRT 171 words and MFrCoE 4,938, these sets remain only a subset of actual Quebec usage. 
Regional or highly-specialized expressions may be missing, biasing model evaluation against cases not covered by our corpus \citep{faisal-etal-2024-dialectbench}.

\paragraph{Evaluation Scope}
Our evaluation of QFrCoRE, QFrCoRT and MFrCoE is conducted exclusively in a zero-shot setting, which highlights the out-of-the-box understanding of idioms by pretrained models. While this offers a clear view of initial model capabilities, it omits popular performance improvement strategies, such as in-context learning through few-shot examples, continual pre-training, or fine-tuning. A complete picture of a model’s utility requires evaluation across different adaptation strategies, not just a single point of assessment \citep{liang2022holistic}.

\paragraph{Definitional Understanding vs. Pragmatic Appropriateness}
Another limitation is that our multiple-choice format evaluates \textit{definitional understanding} but not \textit{pragmatic appropriateness}. 
A model might correctly identify that the expression \enquote{\textit{lâcher son fou}} means to let loose and have fun, but it has no capacity to understand the social context in which the expression is suitable (e.g. among friends on a Friday night) versus where it would be inappropriate (e.g. in a formal business meeting). 
This is especially critical in Quebecois French, where the use of certain terms, particularly those related to \textit{sacres} (religious-based swear words), is heavily dependent on social register and context. 
Consequently, a model could achieve a perfect score on our benchmarks and still fail at generating socially-aware and appropriate Quebecois dialogue, as our evaluation does not capture this crucial layer of linguistic competence.

\paragraph{Potential for Data Contamination from Online Sources}
A limitation of this article is the risk of data leakage from the evaluation corpora into the LLMs' training data. 
The corpora were constructed using publicly-available online sources, including the Canada-Media portal and pages from McGill University and Québec-Cité. 
Given that many of the \llm{} benchmarked LLMs are trained on vast web scrapes, it is probable that content from these specific websites was included in their training sets. 
Consequently, a model's ability to provide a correct definition for an idiom might not stem from genuine understanding but from its ability to recall information it memorized during training \cite{yang2023rethinking,xu2024benchmark}. 
This data contamination would lead to an overestimation of a model's true capabilities.

\paragraph{Hardware-Imposed Constraints on Model Selection}

The study’s findings on open-source models are constrained by the available hardware. 
The experiments were run on three NVIDIA GPUs (see \autoref{an:hardware} for details), which limited the evaluation to models up to approximately 32 billion parameters. 
This practical constraint meant that larger, and often more powerful \cite{kaplan2020scaling}, open-source models (e.g. those with 70B+ parameters) could not be included in the benchmark. 
As a result, the paper's conclusions may not be fully representative of the entire open-source landscape, as the most capable models from the community were omitted from this analysis.

\section*{Ethical Considerations}
The development and release of the QFrCoRE and QFrCoRT datasets, as with any corpus designed to advance LM capabilities, carry ethical implications that warrant careful consideration.

\paragraph{Intended Use and Dual Nature of LLMs}
Our primary goal in creating QFrCoRE and QFrCoRT is to equip the Quebec-French NLP community with reliable benchmarks for measuring progress in NLU, LLM language competency, and idiom comprehension in their language. 
However, we acknowledge that improvements driven by these datasets also fuel the development of ever more powerful LLMs. Such models possess a dual-use character: they can enable valuable applications like enhanced region-specific translation or educational tools. However, they may also be misused to generate persuasive disinformation, automate social manipulation, or produce harmful content at scale \citep{bender2021dangers}.

\paragraph{Mitigation and Positive Impact}
Despite these risks, releasing public benchmarks such as QFrCoRE and QFrCoRT is crucial for promoting transparency and accountability in AI. By providing two complementary datasets focused on the distinctive idioms of Quebec-French, QFrCoRE and QFrCoRT empower researchers and practitioners to assess and compare model performance on region-specific language phenomena rigorously. Moreover, making these resources openly available encourages the community to identify and red-team potential harms such as dialectal bias or misuse of idiomatic mappings and to develop mitigation strategies in line with best practices for reducing LLM harms \citep{ganguli2022red}.

\paragraph{Data Provenance}
QFrCoRE and QFrCoRT are built from publicly-available, printed lexicographic sources (e.g. dictionaries and reference works). 
Although these materials are published for human readers and educational use, their authors and editors did not explicitly consent to downstream use in training or evaluating large-scale AI systems.
Repurposing such content for evaluation, therefore, raises questions of ownership and licensing.

\paragraph{Representational Harms and Stereotyping}
An ethical challenge in creating any dialect-specific corpus lies in the risk of representational harm. 
The act of selecting which expressions and terms to include is an act of curation that can inadvertently shape how a linguistic community is perceived by AI systems. 
While we aimed for broad coverage, our corpora might unintentionally over-represent certain types of idioms—such as folksy, archaic, or highly informal expressions, at the expense of others.
Models trained or evaluated on such a dataset could learn to generate text that caricatures Quebecois speakers, reducing a vibrant and diverse dialect to a set of stereotypes. 
This could lead to downstream applications that perpetuate harmful clichés, for instance, by making chatbots interacting with Quebecers sound like exaggerated, folksy caricatures.
Acknowledging this, we recognize that the curation of dialectal data carries a profound responsibility to represent the community authentically and avoid reinforcing stereotypes.

\section*{Acknowledgements}
This research was made possible thanks to the support of a Canadian insurance company, NSERC research grant RDCPJ 537198-18 and FRQNT doctoral research grant. We thank the reviewers for their comments regarding our work.

\bibliography{anthology,custom}
\bibliographystyle{acl_natbib}

\appendix

\section{Online Source Details}
\label{an:sources}
We present in this section the web scraped online source details:

\begin{itemize}[leftmargin=*, noitemsep, topsep=0ex]
    \item \textbf{\href{https://canada-media.ca/expressions-quebecoises/}{Canada-Media}} is a comprehensive guide to 250 common Quebecois expressions. 
    It also introduces how these unique phrases reflect the spirit and humour of the people of Quebec. 
    The guide suggests that understanding these expressions is beneficial for tourists, new residents, and anyone interested in the local culture.
    \item \textbf{\href{https://vivreenfrancais.mcgill.ca/capsules-linguistiques/expressions-quebecoises/}{Vivre En Francais from McGill University (Expressions)}} is a glossary of common Quebecois expressions. 
    It is organized alphabetically and provides definitions, examples, and sometimes cultural context, such as connections to hockey or the seasons. 
    The page aims to help people, especially those in the McGill community, understand the unique French spoken in Quebec.
    \item \textbf{\href{https://vivreenfrancais.mcgill.ca/capsules-linguistiques/anglicismes/}{Vivre En Francais from McGill University (Anglicismes)}} is also a glossary that focuses on anglicisms, which are words or phrases borrowed from English. 
    It provides a list of common anglicisms to avoid in French and offers the correct French alternatives. The page is structured alphabetically and includes examples of correct and incorrect usage to help French speakers refine their language skills.
    \item \textbf{\href{https://vivreengaspesie.com/15-expressions-typiquement-gaspesiennes/}{Vivre en Gaspésie}} is a website that celebrates the Gaspé Peninsula's unique dialect by presenting 15 typical expressions from the region. 
    For each expression, the website provides the   pronunciation, meaning, and often its origin. Many expressions are linked to specific parts of the Gaspé, and the article highlights the colourful and imaginative language of the region.
    \item \textbf{\href{https://www.quebec-cite.com/fr/ville-quebec/expressions-quebecoises}{Québec-Cité}} is an official website of the City of Quebec that offers a practical guide to 100 common Quebecois expressions for travellers. 
    The expressions are categorized by theme, making it easy to find relevant phrases for various situations. 
\end{itemize}

\input{annexe_model}

\section{Hardware and Private LLM Inference Budget}
\label{an:hardware}
\subsection{Hardware}
We rely on three NVIDIA RTX 6000 ADA with 49 GB of memory, without memory pooling, thus the maximum size we can fit is around 32B parameters in order to have a sufficient batch size to process the experiment in a reasonable timeframe (approximatively a month).

\subsection{Private LLM Inference Budget}
We allocated a budget of approximately 750 USD for using private LLM APIs (e.g. OpenAI, Anthropic) during development, prototyping, and adjusting our prompts. For the complete inference loop for all selected private LLMs for all the tasks, we spent a budget of nearly 2,500 USD.

\section{Detailed Results}
\label{an:results}
The complete results of all \llm{} LLMs tested on QFrCoRT, QFrCoRE and MFrCoE are presented in \autoref{tab:results}.
\begin{table*}[ht!]
  \centering
  \begin{adjustbox}{max width=\textwidth, max totalheight=0.98\textheight, keepaspectratio}
    \begin{tabular}{lccccRRRlccccRRR}
      \toprule
      \textbf{LLM} & OS & Re & Fr & It & \multicolumn{1}{c}{\begin{tabular}[c]{@{}c@{}}\textbf{QFrCoRT}\\ Acc. (\%)\end{tabular}} & \multicolumn{1}{c}{\begin{tabular}[c]{@{}c@{}}\textbf{QFrCoRE}\\ Acc. (\%)\end{tabular}} & \multicolumn{1}{c}{\begin{tabular}[c]{@{}c@{}}\textbf{MFrCoE}\\ Acc. (\%)\end{tabular}} & \textbf{LLM} & OS & Re & Fr & It & \multicolumn{1}{c}{\begin{tabular}[c]{@{}c@{}}\textbf{QFrCoRT}\\ Acc. (\%)\end{tabular}} & \multicolumn{1}{c}{\begin{tabular}[c]{@{}c@{}}\textbf{QFrCoRE}\\ Acc. (\%)\end{tabular}} & \multicolumn{1}{c}{\begin{tabular}[c]{@{}c@{}}\textbf{MFrCoE}\\ Acc. (\%)\end{tabular}}\\
      \midrule
    \texttt{Apertus-8B-2509} & X &  &  &  & 14.04 & 9.76 & 14.50 & \texttt{Kimi-k2-0905} & X &  &  &  & 85.38 & 82.82 & 82.83 \\
    \texttt{Apertus-8B-it-2509} & X &  &  & X & 18.13 & 13.40 & 12.62 & \texttt{Kimi-k2-thinking} & X & X &  &  & 88.45& 85.65 & 84.39 \\
    \texttt{Aya-23-8b} & X &  &  &  & 12.87 & 9.99 & 9.86 & \texttt{Llama-3.2-1B} & X & X &  &  & 11.11 & 9.67 & 10.98 \\
    \texttt{Aya-expanse-8b} & X &  &  &  & 7.60 & 9.37 & 0.00 & \texttt{Llama-3.2-1B-it} & X & X &  & X & 12.28 & 15.93 & 12.60 \\
    \texttt{C4ai-aya-expanse-32b} & X &  &  &  & 74.85 & 53.87 & 81.53 & \texttt{Llama-3.2-3B} & X & X &  &  & 12.87 & 9.67 & 16.48 \\
    \texttt{C4ai-aya-expanse-8b} & X &  &  &  & 54.39 & 34.08 & 67.33 & \texttt{Llama-3.2-3B-it} & X & X &  & X & 24.56 & 17.61 & 36.29 \\
    \texttt{Chocolatine-14B-it-DPO-v1.3} & X &  & X & X & 10.53 & 9.61 & 33.19 & \texttt{Lucie-7B} & X &  & X &  & 7.60 & 10.12 & 11.14 \\
    \texttt{Chocolatine-2-14B-it-v2.0.3} & X &  & X & X & 12.28 & 11.33 & 72.09 & \texttt{Lucie-7B-it-human-data} & X &  & X & X & 10.53 & 9.91 & 10.02 \\
    \texttt{Claude-haiku-4-5-20251001} &  &  &  &  & 92.98 & 87.29 & 97.14 & \texttt{Lucie-7B-it-v1.1} & X &  & X & X & 17.54 & 17.35 & 13.63 \\
    \texttt{Claude-opus-4-1-20250805} &  &  &  &  & 99.42 & 95.38 & 99.01 & \texttt{Meta-Llama-3.1-8B} & X & X &  &  & 9.36 & 9.78 & 11.20 \\
    \texttt{Claude-opus-4-20250514} &  & X &  &  & 97.66 & 93.46 & 98.95 & \texttt{Meta-Llama-3.1-8B-it} & X & X &  & X & 9.36 & 20.14 & 31.75 \\
    \texttt{Claude-sonnet-4-20250514} &  & X &  &  & 97.66 & 91.75 & 98.46 & \texttt{Mistral-large-latest} &  & X &  &  & 90.64 & 84.03 & 96.56 \\
    \texttt{Claude-sonnet-4-5-20250929} &  &  &  &  & 97.66 & 93.40 & 98.54 & \texttt{Mixtral-8x7B-it-v0.1} & X &  &  & X & 11.70 & 9.86 & 5.18 \\
    \texttt{Command-a-03-2025} & X &  &  &  & 92.40 & 82.54 & 95.85 & \texttt{Mixtral-8x7B-v0.1} & X &  &  &  & 11.70 & 9.43 & 8.04 \\
    \texttt{Command-r-08-2024} & X &  &  &  & 15.20 & 14.70 & 35.64 & \texttt{o1-2024-12-17} &  & X &  &  & 95.91 & 85.41 & 97.81 \\
    \texttt{Command-r-plus-08-2024} & X &  &  &  & 67.84 & 44.57 & 71.85 & \texttt{o1-mini-2024-09-12} &  & X &  &  & 77.78 & 70.49 & 79.34 \\
    \texttt{Command-r7b-12-2024} & X &  &  &  & 40.35 & 41.27 & 69.89 & \texttt{o3-2025-04-16} &  & X &  &  & 95.91 & 86.01 & 97.81 \\
    \texttt{CroissantLLMBase} & X &  & X &  & 9.36 & 10.58 & 9.68 & \texttt{o3-mini-2025-01-31} &  & X &  &  & 87.72 & 76.97 & 93.64 \\
    \texttt{DeepSeek-chat} &  &  &  &  & 92.40 & 83.92 & 96.42 & \texttt{OLMo-2-0325-32B} & X &  &  &  & 5.85 & 8.29 & 7.51 \\
    \texttt{DeepSeek-R1-0528-Qwen3-8B} & X &  &  &  & 6.43 & 3.93 & 1.42 & \texttt{OLMo-2-0325-32B-it} & X &  &  & X & 2.92 & 3.95 & 1.70 \\
    \texttt{DeepSeek-R1-Distill-Llama-8B} & X & X &  &  & 11.11 & 10.49 & 9.82 & \texttt{OLMo-2-0425-1B} & X &  &  &  & 12.87 & 9.48 & 10.13 \\
    \texttt{DeepSeek-R1-Distill-Qwen-14B} & X & X &  &  & 35.67 & 32.18 & 55.97 & \texttt{OLMo-2-0425-1B-it} & X &  &  & X & 32.16 & 12.82 & 19.52 \\
    \texttt{DeepSeek-R1-Distill-Qwen-32B} & X & X &  &  & 53.22 & 31.04 & 57.76 & \texttt{OLMo-2-1124-13B} & X &  &  &  & 13.45 & 9.56 & 10.49 \\
    \texttt{DeepSeek-R1-Distill-Qwen-7B} & X & X &  &  & 12.87 & 8.72 & 9.48 & \texttt{OLMo-2-1124-13B-it} & X &  &  & X & 12.87 & 12.30 & 15.78 \\
    \texttt{DeepSeek-reasoner} &  & X &  &  & 91.23 & 84.98 & 95.65 & \texttt{OLMo-2-1124-7B} & X &  &  &  & 9.94 & 9.76 & 11.38 \\
    \texttt{Deepthink-Reasoning-14B} & X & X &  &  & 31.58 & 32.87 & 44.07 & \texttt{OLMo-2-1124-7B-it} & X &  &  & X & 8.19 & 9.26 & 12.70 \\
    \texttt{Deepthink-Reasoning-7B} & X & X &  &  & 13.45 & 15.73 & 22.56 & \texttt{Phi-3.5-mini-it} & X &  &  & X & 12.28 & 5.42 & 5.55 \\
    \texttt{French-Alpaca-Llama3-8B-it-v1.0} & X & X & X & X & 16.37 & 23.74 & 48.26 & \texttt{Phi-4} & X &  &  &  & 36.84 & 18.33 & 33.05 \\
    \texttt{Gemini-2.5-flash} &  &  &  &  & 95.91 & 87.80 & 97.61 & \texttt{Pixtral-large-latest} & X &  &  &  & 86.55 & 77.25 & 94.78 \\
    \texttt{Gemini-2.5-pro} &  & X &  &  & 96.49 & 90.68 & 98.58 & \texttt{Qwen-max} &  &  &  &  & 70.76 & 67.80 & 75.70 \\
    \texttt{Gemini-3-pro-preview} &  &  &  &  & 85.07 & 80.71 & 99.13 & \texttt{Qwen2.5-0.5B} & X &  &  &  & 11.11 & 9.41 & 11.32 \\
    \texttt{Gemma-2-27b} & X & X &  &  & 4.68 & 6.15 & 4.09 & \texttt{Qwen2.5-0.5B-it} & X &  &  & X & 8.19 & 10.99 & 11.10 \\
    \texttt{Gemma-2-27b-it} & X & X &  & X & 15.20 & 27.78 & 38.62 & \texttt{Qwen2.5-1.5B} & X &  &  &  & 16.37 & 13.73 & 20.94 \\
    \texttt{Gemma-2-2b} & X & X &  &  & 12.87 & 10.64 & 11.75 & \texttt{Qwen2.5-1.5B-it} & X &  &  & X & 18.13 & 14.94 & 22.78 \\
    \texttt{Gemma-2-2b-it} & X & X &  & X & 5.26 & 12.84 & 9.68 & \texttt{Qwen2.5-14B} & X &  &  &  & 45.61 & 34.79 & 48.99 \\
    \texttt{Gemma-2-9b} & X & X &  &  & 5.85 & 2.18 & 0.43 & \texttt{Qwen2.5-14B-it} & X &  &  & X & 29.82 & 33.13 & 43.76 \\
    \texttt{Gemma-2-9b-it} & X & X &  & X & 8.19 & 6.76 & 5.16 & \texttt{Qwen2.5-32B} & X &  &  &  & 53.22 & 48.50 & 72.24 \\
    \texttt{Glm-4.5} & X & X &  &  & 87.72 & 82.99 & 96.17 & \texttt{Qwen2.5-32B-it} & X &  &  & X & 38.60 & 39.02 & 50.73 \\
    \texttt{GPT-4.1-2025-04-14} &  &  &  &  & 94.15 & 86.73 & 97.31 & \texttt{Qwen2.5-3B} & X &  &  &  & 19.88 & 15.43 & 16.12 \\
    \texttt{GPT-4.1-mini-2025-04-14} &  &  &  &  & 92.40 & 82.80 & 96.13 & \texttt{Qwen2.5-3B-it} & X &  &  & X & 12.87 & 11.74 & 9.72 \\
    \texttt{GPT-4o-2024-08-06} &  &  &  &  & 93.57 & 83.94 & 96.35 & \texttt{Qwen2.5-7B} & X &  &  &  & 23.39 & 13.10 & 38.56 \\
    \texttt{GPT-4o-mini-2024-07-18} &  &  &  &  & 90.64 & 73.88 & 92.57 & \texttt{Qwen2.5-7B-it} & X &  &  & X & 15.20 & 15.95 & 22.86 \\
    \texttt{GPT-5-2025-08-07} &  & X &  &  & 95.32 & 87.93 & 97.59 & \texttt{Qwen3-14B} & X &  &  &  & 28.07 & 33.48 & 53.95 \\
    \texttt{GPT-5-mini-2025-08-07} &  & X &  &  & 92.40 & 77.36 & 94.53 & \texttt{Qwen3-14B-Base} & X &  &  &  & 50.88 & 57.39 & 84.71 \\
    \texttt{GPT-5.1-2025-11-13} &  &  &  &  & 95.91 & 90.09 & 98.30 & \texttt{Qwen3-235b-a22b} & X &  &  &  & 79.53 & 75.74 & 91.86 \\
    \texttt{GPT-oss-120b} &  & X &  &  & 76.61 & 66.52 & 85.68 & \texttt{Qwen3-235b-a22b-thinking-2507} & X & X &  &  & 70.18 & 64.23 & 55.67 \\
    \texttt{GPT-oss-20b} & X & X &  &  & 8.77 & 6.65 & 53.08 & \texttt{QwQ-32B} & X & X &  &  & 22.81 & 24.45 & 47.25 \\
    \texttt{Granite-3.2-8b-it} & X &  &  & X & 10.53 & 11.33 & 14.32 & \texttt{Random Selection} & - & - & - & - & 12.28 & 9.93 & 10.41 \\
    \texttt{Granite-3.3-8b-base} & X &  &  &  & 12.87 & 9.82 & 10.43 & \texttt{Reka-flash-3} & X & X &  &  & 11.70 & 10.32 & 10.23 \\
    \texttt{Granite-3.3-8b-it} & X &  &  & X & 17.54 & 12.67 & 32.87 & \texttt{S1.1-32B} & X & X &  &  & 56.14 & 47.20 & 69.79 \\
    \texttt{Grok-3-fast-latest} &  & X &  &  & 92.40 & 79.43 & 95.85 & \texttt{SmolLM2-1.7B} & X &  &  &  & 9.94 & 9.37 & 10.59 \\
    \texttt{Grok-3-latest} &  & X &  &  & 91.23 & 79.47 & 95.93 & \texttt{SmolLM2-1.7B-it} & X &  &  & X & 11.11 & 9.76 & 23.11 \\
    \texttt{Grok-3-mini-fast-latest} &  & X &  &  & 88.89 & 82.54 & 95.18 & \texttt{SmolLM2-135M} & X &  &  &  & 9.36 & 10.25 & 9.88 \\
    \texttt{Grok-3-mini-latest} &  & X &  &  & 90.64 & 82.50 & 95.40 & \texttt{SmolLM2-135M-it} & X &  &  & X & 9.36 & 9.37 & 9.78 \\
    \texttt{Grok-4-0709} &  &  &  &  & 95.32 & 85.84 & 97.35 & \texttt{SmolLM2-360M} & X &  &  &  & 7.60 & 11.37 & 9.64 \\
    \texttt{Grok-4-fast-non-reasoning} &  &  &  &  & 90.64 & 84.80 & 96.74 & \texttt{SmolLM2-360M-it} & X &  &  & X & 12.87 & 9.41 & 10.04 \\
    \texttt{Grok-4-fast-reasoning} &  & X &  &  & 89.47 & 83.14 & 95.91 & & &  \\
      \bottomrule
    \end{tabular}
  \end{adjustbox}
  \caption{Performance of all \llm{} models on QFrCoRT, QFrCoRE and MFrCoE tasks in alphabetic order. \enquote{OS} indicates open-source models, \enquote{Re} indicates reasoning models,  \enquote{Fr} indicates French-language models, and \enquote{It} indicates variant models with instruction tuning. Scores are accuracy (Acc.) (\%).}
  \label{tab:results}
\end{table*}
\end{document}

%% file: annexe_model.tex
\section{Selected LLM Details}
\label{an:selectedllmdetails}
We present in \autoref{tab:selectedllm} the comprehensive suite of open-source LLMs we could fit on our hardware (see \autoref{an:hardware}) or can be process on a provider API (e.g. Mistral AI) or a third party service (e.g. \href{https://openrouter.ai/}{OpenRouter}) (details in \autoref{tab:selectedllm}), detailing their origins and respective sizes, while in \autoref{tab:selectedprivatellm}, we present the comprehensive suite of private LLMs benchmarked in our study.
The selection was curated to cover a wide spectrum of parameter counts, and to include those with specializations in French ($\Upsilon$) or reasoning ($\Gamma$).
All LLMs are downloaded from the \href{https://huggingface.co/models}{HuggingFace Model repository} \citep{wolf2020huggingfaces} using default parameters.

\begin{table*}
    \centering
    \caption{The selected open-source LLM used in our work, along with their source and size. \enquote{$\Upsilon$} are model that have a specialization in French, while \enquote{$\Gamma$} are model marketed as reasoning LLM. Model with an \enquote{*} have either been run on a third party provider (e.g. \href{https://openrouter.ai/}{OpenRouter}) or the provided API service (e.g. Mistral AI).}
    \label{tab:selectedllm}
    \resizebox{\textwidth}{!}{%
    \begin{tabular}{llcllc}
        \toprule
        LLM & Source & Size&        LLM & Source & Size\\
\midrule
\texttt{Apertus-8B-2509} & \citet{swissai2025apertus} & 8B & \texttt{Lucie-7b-it} ($\Upsilon$) & \citet{openllm2025lucie} & 6.71B \\
\texttt{Apertus-8B-it-2509} & \citet{swissai2025apertus} & 8B & \texttt{Lucie-7b} ($\Upsilon$) & \citet{openllm2025lucie} & 6.71B \\
\texttt{Aya-23-8b} & \citet{aryabumi2024aya} & 8B & \texttt{Meta-Llama-$3.1$-8b-it} ($\Gamma$) & \citet{grattafiori2024llama} & 8B \\
\texttt{Aya-expanse-32b} & \citet{dang2024ayaexpansecombiningresearch} & 32B & \texttt{Meta-Llama-$3.1$-8b} ($\Gamma$) & \citet{grattafiori2024llama} & 8B \\
\texttt{Aya-expanse-8b} & \citet{dang2024ayaexpansecombiningresearch} & 8B & \texttt{Mistral-large-latest}* 
(v2) ($\Gamma$) &  N/A & 675B \\
\texttt{Chocolatine-14b-it} ($\Upsilon$) & \citet{chocolatine} & 14B & \texttt{Mixtral-8x7b-it} & \citet{rastogi2025magistral} & 46.7B \\
\texttt{Chocolatine-2-14b-it} ($\Upsilon$) & \citet{chocolatinev2} & 14.8B & \texttt{Mixtral-8x7b} & \citet{rastogi2025magistral} & 46.7B \\
\texttt{Command-a-03-2025}* & \citet{cohere2025commandaenterprisereadylarge} & 111B & \texttt{OLMo-2-13B-it} & \citet{olmo20242olmo2furious} & 13.7B \\
\texttt{Command-a-reasoning-08-2025}* ($\Gamma$) & \citet{cohere2025commandaenterprisereadylarge} & 111B & \texttt{OLMo-2-13B} & \citet{olmo20242olmo2furious} & 13.7B \\
\texttt{Command-r-08-2024} & \citet{cohere2025commandaenterprisereadylarge} & 32B & \texttt{OLMo-2-1B-it} & \citet{olmo20242olmo2furious} & 1.48B \\
\texttt{Command-r-plus-08-2024}* & \citet{cohere2025commandaenterprisereadylarge} & 104B & \texttt{OLMo-2-1B} & \citet{olmo20242olmo2furious} & 1.48B \\
\texttt{Command-r7b-12-2024} & \citet{cohere2025commandaenterprisereadylarge} & 8B & \texttt{OLMo-2-32B-it} & \citet{olmo20242olmo2furious} & 32.2B \\
\texttt{CroissantLLM-Base} ($\Upsilon$) & \citet{faysse2024croissantllm} & 1.3B & \texttt{OLMo-2-32B} & \citet{olmo20242olmo2furious} & 32.2B \\
\texttt{DeepSeek-R1-distill-Llama-8b} ($\Gamma$) & \citet{deepseekai2025deepseekr1incentivizingreasoningcapability} & 8.03B & \texttt{OLMo-2-7B-it} & \citet{olmo20242olmo2furious} & 7.3B \\
\texttt{DeepSeek-R1-distill-Qwen-14b} ($\Gamma$) & \citet{deepseekai2025deepseekr1incentivizingreasoningcapability} & 14.8B & \texttt{OLMo-2-7B} & \citet{olmo20242olmo2furious} & 7.3B \\
\texttt{DeepSeek-R1-distill-Qwen-32b} ($\Gamma$) & \citet{deepseekai2025deepseekr1incentivizingreasoningcapability} & 32.8B & \texttt{Phi-$3.5$-mini-it} & \citet{abdin2024phi3technicalreporthighly} & 3.8B \\
\texttt{DeepSeek-R1-distill-Qwen-7b} ($\Gamma$) & \citet{deepseekai2025deepseekr1incentivizingreasoningcapability} & 7.62B & \texttt{Phi-4} & \citet{abdin2024phi} & 14.7B \\
\texttt{DeepSeek-R1-distill-Qwen3-8b} ($\Gamma$) & \citet{deepseekai2025deepseekr1incentivizingreasoningcapability} & 5.27B & \texttt{Pixtral-large-latest} & N/A & 123B \\
\texttt{DeepSeek-chat} & \citet{liu2024deepseek} & 236B & \texttt{Qwen$2.5$-$1.5$b} & \citet{hui2024qwen2} & 1.5B \\
\texttt{DeepSeek-reasoner} ($\Gamma$) & \citet{liu2024deepseek} & 236B & \texttt{Qwen$2.5$-14b-it} & \citet{hui2024qwen2} & 14.7B \\
\texttt{Deepthink-reasoning-14b} ($\Gamma$) & \citet{deepthink2} & 14.8B & \texttt{Qwen$2.5$-14b} & \citet{hui2024qwen2} & 14.7B \\
\texttt{Deepthink-reasoning-7b} ($\Gamma$) & \citet{deepthink1} & 7.62B & \texttt{Qwen$2.5$-32b-it} & \citet{hui2024qwen2} & 32.8B \\
\texttt{French-Alpaca-Llama3-8b-it} ($\Upsilon$, $\Gamma$)   & \citet{alpaca} & 8.03B & \texttt{Qwen$2.5$-32b} & \citet{hui2024qwen2} & 32.8B \\
\texttt{GLM-$4.5$}* ($\Gamma$) & \citet{zeng2025glm} & 358B & \texttt{Qwen$2.5$-3b-it} & \citet{hui2024qwen2} & 3B \\
\texttt{GPT-oss-120B}* ($\Gamma$) & \citet{gptoss} & 120B & \texttt{Qwen$2.5$-3b} & \citet{hui2024qwen2} & 3B \\
\texttt{GPT-oss-20b} ($\Gamma$) & \citet{gptoss} & 21.5B & \texttt{Qwen$2.5$-7b-it} & \citet{hui2024qwen2} & 7.6B \\
\texttt{Gemma-2-27b-it} ($\Gamma$) & \citet{mesnard2024gemma} & 27.2B & \texttt{Qwen$2.5$-7b} & \citet{hui2024qwen2} & 7.6B \\
\texttt{Gemma-2-27b} ($\Gamma$) & \citet{mesnard2024gemma} & 27.2B & \texttt{Qwen3-14b-base} & \citet{qwen3technicalreport} & 14.8B \\
\texttt{Gemma-2-2b-it} ($\Gamma$) & \citet{mesnard2024gemma} & 27.2B & \texttt{Qwen3-14b} & \citet{qwen3technicalreport} & 8.76B \\
\texttt{Gemma-2-2b} ($\Gamma$) & \citet{mesnard2024gemma} & 2.6B & \texttt{Qwen3-235b-a22b-thinking-2507}* ($\Gamma$) & \citet{qwen3technicalreport} & 235B \\
\texttt{Gemma-2-9b-it} ($\Gamma$) & \citet{mesnard2024gemma} & 9B & \texttt{Qwen3-235b-a22b}* & \citet{qwen3technicalreport} & 235B \\
\texttt{Gemma-2-9b} ($\Gamma$) & \citet{mesnard2024gemma} & 9.2B & \texttt{Reka-flash-3} ($\Gamma$) & \citet{reka} & 20.9B \\
\texttt{Granite$3.2$-8B} & \citet{granite2024granite} & 8.17B & \texttt{S1.1-32b} ($\Gamma$) & \citet{s11} & 32.8B \\
\texttt{Granite$3.3$-8B-base} & \citet{granite2024granite} & 8.17B & \texttt{SmolLM2-$1.7$b-it} & \citet{allal2025smollm2smolgoesbig} & 1.7B \\
\texttt{Granite$3.3$-8B-it} & \citet{granite2024granite} & 8.17B & \texttt{SmolLM2-$1.7$b} & \citet{allal2025smollm2smolgoesbig} & 1.7B \\
\texttt{Llama-$3.2$-1b-it} ($\Gamma$) & \citet{grattafiori2024llama} & 1.2B & \texttt{SmolLM2-135m-it} & \citet{allal2025smollm2smolgoesbig} & 134.5M \\
\texttt{Llama-$3.2$-1b} ($\Gamma$) & \citet{grattafiori2024llama} & 1.2B & \texttt{SmolLM2-135m} & \citet{allal2025smollm2smolgoesbig} & 134.5M \\
\texttt{Llama-$3.2$-3b-it} ($\Gamma$) & \citet{grattafiori2024llama} & 3.21B & \texttt{SmolLM2-360m-it} & \citet{allal2025smollm2smolgoesbig} & 361.8M \\
\texttt{Llama-$3.2$-3b} ($\Gamma$) & \citet{grattafiori2024llama} & 3.21B & \texttt{SmolLM2-360m} & \citet{allal2025smollm2smolgoesbig} & 361.8M \\
\texttt{Lucie-7b-it-human-data} ($\Upsilon$) & \citet{openllm2025lucie} & 6.71B &  &  &  \\
\bottomrule
    \end{tabular}%
}
\end{table*}

\begin{table*}
    \centering
    \caption{The selected private LLMs used in our work, along with their source. \enquote{$\Gamma$} indicates models marketed as reasoning LLMs.}
    \label{tab:selectedprivatellm}
    \resizebox{\textwidth}{!}{%
    \begin{tabular}{llll}
        \toprule
        \textbf{LLM} & \textbf{Source} & \textbf{LLM} & \textbf{Source} \\
        \cmidrule{1-2} \cmidrule{3-4}
        \texttt{Claude-Haiku-4-5-20251001} ($\Gamma$) & Anthropic & \texttt{GPT-5-mini-2025-08-07} ($\Gamma$) & OpenAI \\
        \texttt{Claude-Opus-4-1-20250805} ($\Gamma$) & Anthropic & \texttt{GPT-5.1} ($\Gamma$) & OpenAI \\
        \texttt{Claude-Opus-4-20250514} ($\Gamma$) & Anthropic & \texttt{Grok-3-fast-latest} ($\Gamma$) & xAI \\
        \texttt{Claude-Sonnet-4-20250514} ($\Gamma$) & Anthropic & \texttt{Grok-3-latest} ($\Gamma$) & xAI \\
        \texttt{Claude-Sonnet-4-5-20250929} ($\Gamma$) & Anthropic & \texttt{Grok-3-mini-fast-latest} ($\Gamma$) & xAI \\
        \texttt{Gemini-$2.5$-flash} & Google & \texttt{Grok-3-mini-latest} ($\Gamma$) & xAI \\
        \texttt{Gemini-$2.5$-pro} ($\Gamma$) & Google & \texttt{Grok-4-0709} ($\Gamma$) & xAI \\
        \texttt{Gemini-3-pro} ($\Gamma$) & Google & \texttt{Grok-4-fast-non-reasoning} ($\Gamma$) & xAI \\
        \texttt{GPT-$4.1$-2025-04-14} & OpenAI & \texttt{Grok-4-fast-reasoning} ($\Gamma$) & xAI \\
        \texttt{GPT-$4.1$-mini-2025-04-14} & OpenAI & \texttt{o1-2024-12-17} ($\Gamma$) & OpenAI \\
        \texttt{GPT-4o-2024-08-06} & OpenAI & \texttt{o1-mini-2024-09-12} ($\Gamma$) & OpenAI \\
        \texttt{GPT-4o-mini-2024-07-18} & OpenAI & \texttt{o3-2025-04-16} ($\Gamma$) & OpenAI \\
        \texttt{GPT-5-2025-08-07} ($\Gamma$) & OpenAI & \texttt{o3-mini-2025-01-31} ($\Gamma$) & OpenAI \\
        \bottomrule
    \end{tabular}
}
\end{table*}